\documentclass{article}

\PassOptionsToPackage{numbers, compress}{natbib}
\usepackage[preprint]{neurips_2021}

\usepackage{amsmath}
\usepackage{amssymb}
\usepackage{makecell}
\usepackage{changes}
\usepackage{enumerate}
\usepackage{subfigure}
\usepackage{threeparttable}
\usepackage{epsfig}
\usepackage{diagbox}
\usepackage{bbding}
\usepackage{multirow}

\def\bv #1{\boldsymbol{\rm{#1}}}
\def\bm #1{\boldsymbol{#1}}

\usepackage[utf8]{inputenc} 
\usepackage[T1]{fontenc}    
\usepackage{hyperref}       
\usepackage{url}            
\usepackage{booktabs}       
\usepackage{amsfonts}       
\usepackage{nicefrac}       
\usepackage{microtype}      
\usepackage{xcolor}         

\makeatletter

\DeclareRobustCommand\onedot{\futurelet\@let@token\@onedot}
\def\@onedot{.}
 
\def\ie{\textit{i.e}\onedot}

\def\etal{\textit{et al}\onedot~}

\definecolor{sgreen}{rgb}{0.2,0.6,0.15}
\definecolor{sblue}{rgb}{0,0.3,0.9}

\newcommand{\Tref}[1]{Tab.~\ref{#1}}

\newcommand{\Fref}[1]{Figure~\ref{#1}}

\newcommand{\eref}[1]{Eq.~(\ref{#1})}

\title{Gaze Estimation using Transformer}

\author{Yihua Cheng\\ Beihang University \\yihua\_c@buaa.edu.cn \And Feng Lu\thanks{Corresponding author} \\Beihang University\\lufeng@buaa.edu.cn \\
}

\begin{document}

\maketitle

\begin{abstract}
Recent work has proven the effectiveness of transformers in many computer vision tasks.
However, the performance of transformers in gaze estimation is still unexplored.
In this paper, we employ transformers and assess their effectiveness for gaze estimation. We consider two forms of vision transformer which are pure transformers and hybrid transformers. We first follow the popular ViT and employ a pure transformer to estimate gaze from images. On the other hand, we preserve the convolutional layers and integrate CNNs as well as transformers. The transformer serves as a component to complement CNNs. 
We compare the performance of the two transformers in gaze estimation.
The Hybrid transformer significantly outperforms the pure transformer in all evaluation datasets with less parameters.
We further conduct experiments to assess the effectiveness of the hybrid transformer and explore the advantage of self-attention mechanism. 
Experiments show the hybrid transformer can achieve state-of-the-art performance in all benchmarks with pre-training.
To facilitate further research, we release codes and models in \url{https://github.com/yihuacheng/GazeTR}.
\end{abstract}

\section{Introduction}

Human gaze provides important cues for understanding human cognition~\cite{Rahal_2019_ESP} and behavior~\cite{Dias_2020_WACV}.
It is widely demanded by various fields such as saliency detection~\cite{Wang_2019_TPAMI, Wang_2018_attentionpre}, virtual reality~\cite{Xu_2018_CVPR} and first-person video analysis~\cite{Yu_2020_TPAMI}.
Conventional gaze estimation methods usually build an geometric eye model to calculate human gaze~\cite{Guestrin_2006_TBE}.
Recently, appearance-based gaze estimation with deep learning attracts much attention.
Many convolutional neural network (CNN) based  methods are proposed~\cite{Cheng_2020_tip,Park_2018_ECCV, Cheng_2020_AAAI}.

Appearance-based gaze estimation methods directly learn a mapping function from facial appearance to human gaze~\cite{Zhang_2017_CVPRW}.
However, facial appearance is various due to personal or environmental factors such as head pose and illumination.
These factors are spread and blend in the whole appearance, and complicate the appearance-based gaze estimation problem~\cite{cheng_2021_arxiv}.
This means the learned mapping function should be highly non-linear and have good ability to attend the whole appearance.
Although there are many CNN-based gaze estimation models, more effective gaze estimation models are always crucial and demanded. 

Transformer is proposed by Vaswani~\etal~\cite{vaswani2017attention} and has shown state-of-the-art performance in natural language processing tasks.
Recently, some methods apply transformers into computer vision tasks and achieve excellent performance.
Compared with CNNs, transformers have better ability in capturing global relation.
ViT~\cite{dosovitskiy2020image} uses a pure transformer model in the image classification task and has better results than state-of-the-art convolutional networks.
This progress raises a question, \textbf{\textit{if transformers are suitable for gaze estimation tasks?}}
However, to our knowledge, there is no research about transformers in gaze estimation.
\begin{figure}[t]
	\begin{center}
		\includegraphics[width=0.9\columnwidth]{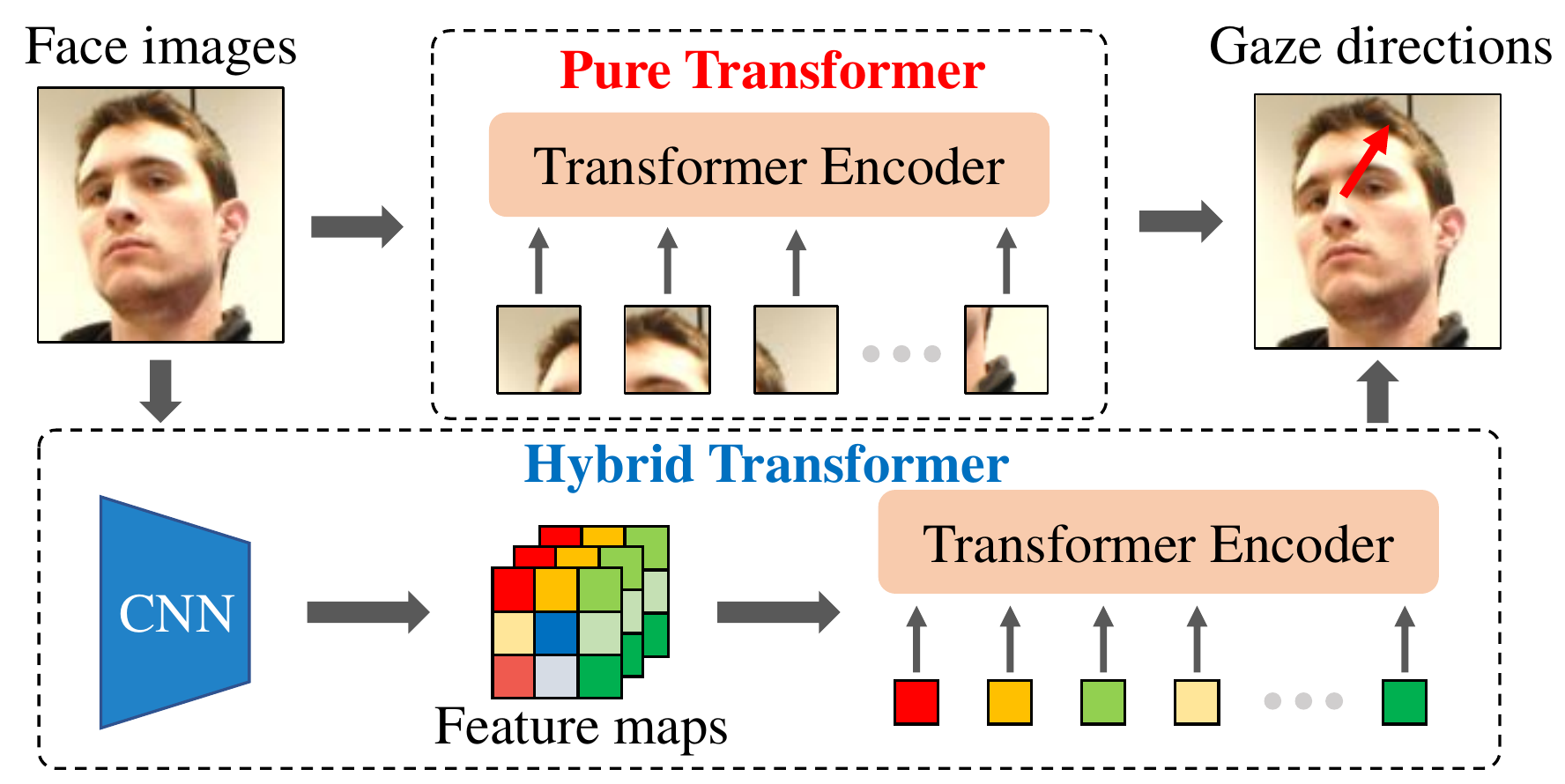}	
	\end{center}
	\caption{We consider two forms of transformers in this paper. The pure transform divides images into multiple patches and estimate gaze from these patches with a transformer encoder. The hybrid transformer first employs a CNN to extract local feature maps and then estimate gaze from the feature maps with a transformer encoder.}
	\label{fig:pretrain}
\end{figure}

In this paper, we employ transformers and assess their effectiveness for gaze estimation.
Inspired by the usage of transformers in other tasks, we consider two different forms of transformers, which are pure transformers and hybrid transformers.
We first directly employ a transformer for gaze estimation,~\ie, pure transformer.
The popular ViT model is applied into gaze estimation.
Detailly, we divide a face image into patches and feed these patches into a transformer encoder for gaze estimation.
On the other hand, we consider the patch division might corrupt the image structure, which might have little influence in classification tasks while greatly impact gaze regression tasks.
Therefore, we further explore the performance of hybrid transformers in gaze estimation.
We use the convolutional layers of ResNet-18 to learn local feature maps from face images, and use a transformer  encode to capture global relations from the feature maps.

Experiments show the pure transformer cannot achieve competitive result in gaze estimation.
However, the hybrid transformer significantly outperforms the pure transformer in many gaze estimation benchmarks.
On the other hand, compared with pure CNNs, \ie, ResNet-18, the hybrid transformer employs an additional transformer encoder which increases $0.2$M (or $1\%$) parameter costs, but achieves significant improvement ranging from $0.18^\circ$ to $0.47^\circ$. 
We believe the hybrid transformer is more suitable for serving as a backbone for future research.
The hybrid transformer also outperform current state-of-the-art methods in popular  benchmarks with pre-training.

Overall, this paper is the first to employ the transformer for gaze estimation. 
The hybrid transformer is proved more suitable for gaze estimation than the pure transformer, and achieves state-of-the art performance in popular gaze estimation benchmarks with pre-training.
We further conduct comprehensive experiments to assess the hybrid transformer.
The impact of convolutional layers, transformer encoders and the self-attention mechanism are respectively evaluated. We also learn the impact of pre-training for transformers in gaze estimation.
Experiments demonstrate integrating transformers can bring larger performance improvement than using pure CNNs.

\section{Related Work}

\subsection{Gaze Estimation}
Gaze estimation with CNNs has significant performance improvement compared with conventional methods~\cite{Zhang_2015_CVPR}.
Recently, many CNN-based gaze estimation methods have been proposed.
Cheng~\etal use a four-stream CNN to estimate gaze from eye images~\cite{Cheng_2018_ECCV}.
They explore the asymmetry between two eyes and propose asymmetric regression.
Park~\etal propose a pictorial representation of eye images~\cite{Park_2018_ECCV}.
They first generate the pictorial representation from eye images and then estimate gaze from the representation.
Chen~\etal use dilated convolutional network to capture the subtle changes in eye images~\cite{Chen_2019_ACCV}.
Wang~\etal align the feature extracted by CNN with adversarial learning~\cite{Wang2_2019_CVPR}.
They also incorporate bayesian inference with the CNN for better performance.
Cheng~\etal propose a coarse-to-fine network to integrate face and eye images~\cite{Cheng_2020_AAAI}.
They estimate a basic gaze from face images and refine the basic gaze with eye images.
These methods all achieve reasonable performance with CNNs.
However, more accurate gaze estimation models are still crucial.

\subsection{Transformers}
Transformer was introduce by Vaswani~\etal and has shown state-of-the-art performance in many natural language processing (NLP) tasks~\cite{vaswani2017attention}.
Vaswani~\etal replace convolutions and recurrent networks with the transformer architecture that contains only self-attention layers, layer normalization and multilayer perceptron layers.
Compared with recurrent networks, the global computations and perfect memory of self-attention layers make transformers more suitable on long sequences.  
Current state-of-the-art methods in NLP are also transformer-based~\cite{devlin2019bert,radford2019language,raffel2020exploring}.

Inspired by the success of transformers in the NLP, transformers also become a hotspot in computer vision filed.
Some works integrate CNNs and transformers, and achieve better performance in object detection and instance segmentation tasks~\cite{carion2020end, chi2020relationnet,zhu2020deformable,sun2020sparse}.
Carion~\etal propose DETR in ~\cite{carion2020end}. 
They consider object detection as a direct set prediction problem, and utilize a transformer encoder-decoder architecture as the detection head.
They first feed images into a CNN for feature extraction, and then input the extracted feature into the detection head for prediction. 
As a result, DETR achieves competitive results compared to Faster R-CNN~\cite{ren2016faster} in quantitative evaluation on COCO~\cite{lin2014microsoft}.

Recently, pure transformer networks show state-of-the-art results in many tasks~\cite{dosovitskiy2020image,kumar2021colorization,chen2021transformer,liu2021swin,arnab2021vivit,han2021transformer}. 
Dosovitskiy~\etal propose Vision Transformer (ViT) in ~\cite{dosovitskiy2020image}.
ViT divides one image into non-overlapping patches.
A pure transformer architecture is directly applied into these patches for image classification.
Dosovitskiy~\etal show ViT can achieve state-of-the-art results with pre-training in large-scale datasets for image classification.
Some works are also proposed for improving the effectiveness of ViT in small datasets~\cite{touvron2020training,pan2021scalable}. 

Although transformers show good ability in many computer vision tasks, the performance of transformer is still unexplored in gaze estimation. In this paper, we employ transformer and assess their effectiveness for gaze estimation.

\section{Gaze Transformers}

Inspired by the usage of transformers in other tasks, we employ two types of transformers for gaze estimation, which are pure transformers and hybrid transformers.

\subsection{Preliminaries}
\label{sec:pre}
We first briefly introduce transformer architecture.
The core of transformers is the self-attention module.
Given a feature matrix $\bm{X}\in\mathbb{R}^{n\times d}$, the feature is projected into queries $\bm{Q}\in\mathbb{R}^{n\times d_k}$, keys $\bm{K}\in\mathbb{R}^{n\times d_k}$ and values $\bm{V}\in\mathbb{R}^{n\times d_v}$ with multilayer perceptron layers (MLP), where $n$ is the batch size, $d$, $d_k$ and $d_v$ is the dimension of each feature.   
The output of self-attention module is computed as:
\begin{equation}
\label{equ:attention}
\rm{Attention}(\bm{Q}, \bm{K}, \bm{V}) = softmax(\frac{\bm{QK}^T}{\sqrt{d_k}})\bm{V}.
\end{equation}

More concretely, transformers contain an encoder and a decoder, where the encoder is widely used in computer vision tasks\footnote{This paper does not relate to transformer decoders, we refer the transformer encoders as transformers in the rest. }. 
The transformer is composed of three components, which are multi-head self-attention (MSA), MLP, and layer normalization (LN)~\cite{ba2016layer}.
MSA extends the self-attention module into multiple subspaces. 
It linearly projects the queries, keys and values $N$ times with different linear projections, where $N$ is the number of heads.
The output values of each head are concatenated and linearly projected to form the final output.
A two-layer MLP is applied between MSA layers for non-linearity, and the LN is used for stable training and faster convergence. Transformers also adopt the skip connection idea~\cite{He_2016_CVPR}.
Overall, we can formulate an one-layer transformer as:

\begin{equation}
\bm{{x}^\prime} = \rm{MSA}(LN(\bm{X})) + \bm{X}
\end{equation}
\begin{equation}
\bm{x} = \rm{MLP}(LN(\bm{{x}^\prime})) + \bm{{x}^\prime}
\end{equation}

where the output $\bm{x}$ usually has the same dimensions as the input $\bm{X}$,~\ie, $\bm{x}\in\mathbb{R}^{n\times d}$. In the other words, it can be stacked for multi-layer transformers.

\subsection{Pure Transformers in Gaze Estimation}
The pure transformer in gaze estimation (GazeTR-Pure) is designed following ViT~\cite{dosovitskiy2020image}.
Given a face image $\bm{I}\in\mathbb{R}^{H\times W\times C}$, $\bm{I}$ is divided into $P\times P$ patches $\bm{I_{i}}\in\mathbb{R}^{\frac{H}{P}\times\frac{W}{P}\times C}$, where $i=1 ... P^2$.
Each patch is flattened and seemed as a feature vector.  
We use a linear projection $T:\mathbb{R}^\frac{HWC}{P^2} \rightarrow \mathbb{R}^D$ to map these feature vectors into $D$-dimensional feature space.
As a result, we acquire the image feature matrix $\bm{z_{img}}\in\mathbb{R}^{P^2\times D}$, where $P^2$ is the length of these feature vectors and $D$ is the dimensions of each feature vector.

We add an extra token $\bm{z_{token}}$ into the image feature matrix. The token is a learnable embedding and has the same dimensions as the feature vector, \ie, $\bm{z_{token}}\in\mathbb{R}^{1\times D}$.  
During training, the token aggregates the feature of other patches with self-attention mechanism,
and finally outputs gaze representations at the output of the transformer.
We also recode the position information of each patches with position embedding. A learnable embedding $\bm{z_{pos}}\in\mathbb{R}^{(1+P^2)\times D}$ is created and added into the image feature matrix.
Overall, we acquire the final feature matrix as 

\begin{equation}
\label{feature}
\bm{z} = [\bm{z_{token}}; \bm{z_{img}}] + \bm{z_{pos}},
\end{equation}

where $[\quad]$ represents concatenation operations and $\bm{z}\in\mathbb{R}^{{(1+P^2)\times D}}$. 

Then, we feed the feature matrix into a transformer.
The transformer processes the feature matrix and outputs a new feature matrix $\bm{z_{out}}\in\mathbb{R}^{{(1+P^2)\times D}}$.
We select the first feature vector (corresponds to the location of  $\bm{z_{token}}$) as gaze representations, and use a MLP to regress gaze from the gaze representation. The whole process can also be formulated as
\begin{equation}
\label{map}
\bv{g} = \rm{MLP}(Transformer(\bm{z})[0,:]),
\end{equation}
where $[0,:]$ represents we select the first row of the feature matrix and $\bv{g}$ is the estimated gaze. The detail of \textit{Transformer} is introduced in the previous section.

\subsection{Hybrid Transformers in Gaze Estimation}
The hybrid transformer in gaze estimation (GazeTR-Hybrid) is composed of a CNN and a transformer. Indeed, we consider the difference between gaze estimation and other tasks such as image classification.
An object is divided into multiple patches and we can also distinguish this object from a local patch.
However, gaze estimation is a regression task and it is hard to predict the human gaze with a local patch such as a half of eye images.
Therefore, we decide to use a CNN to extract local feature from images.
After convolution, each feature contains the information of a local region.
We then feed the feature matrix into a transformer to capture global relations.

More concretely, given a face image $\bm{I}\in\mathbb{R}^{H\times W\times C}$, we first use a CNN to process the face image and acquire feature maps $\bm{f_{img}}\in\mathbb{R}^{h\times w\times c}$.
We then reshape the feature maps into a 2D patch $\bm{f_p}\in\mathbb{R}^{l\times c}$, where $l=h\cdot w$.
It means we get a feature matrix of length $l$ and each feature is $c$-dimensional.
We also add an extra token $\bm{f_{token}}\in\mathbb{R}^{1\times c}$ and a position embedding $\bm{f_{pos}}\in\mathbb{R}^{(l+1)\times c}$ into $\bm{f_p}$ as \eref{feature}.
A transformer and a MLP is used as \eref{map} to regress human gaze.

\subsection{Implementation Details.}
We use $224\times224\times3$ face images for gaze direction estimation.
The estimated gaze is a 2D vector, which contains the yaw and pitch of gaze.
We use L1-loss as the loss function.

For GazeTR-Pure, we divide the image into $14\times14$ patches. The resolution of each patch is $16\times 16$. 
We use a MLP to project the patch into $768$ dimension feature vectors and feed them into a $12$-layer transformer.
We set the number of heads in MSA as $64$ and the hidden size of the two-layer MLP is set as $4096$. A $0.1$ dropout is used after each MLP.

For GazeTR-Hybrid, we employ the convolutional layers of ResNet-18 for feature extraction. The convolutional layers generate $7\times7\times512$ feature maps from face images. We then use an extra $1\times 1$ convolutional layer to scale the channel, and get $7\times 7\times 32$ feature maps.  We feed the feature maps into a 6-layer transformer.
As for transformer, we set the hidden size of two-layer MLP as 512 and perform 8-heads self-attention mechanism. 
The dropout probability is set as 0.1.

\section{Experiments}
\subsection{Setup}
\textbf{Dataset for pre-training.} 
We use ETH-XGaze~\cite{Zhang_2020_ECCV} for pre-training.
ETH-XGaze is collected by high-resolution cameras under indoor environment. 
It contains a total of 1.1M images of 110 subjects.
We use the training set in ETH-XGaze for pre-training which contains 765K images of 80 subjects.
ETH-XGaze provides the normalized data. We directly use the provided data for training.

\textbf{Dataset for evaluation.}
To comprehensive evaluate the model, we select four datasets for evaluation, which are MPIIFaceGaze~\cite{Zhang_2017_CVPRW}, EyeDiap~\cite{Mora_2014_ETRA}, Gaze360~\cite{Kellnhofer_2019_ICCV} and RT-Gene~\cite{Fischer_2018_ECCV}.
We follow ~\cite{cheng2021appearance} to process the first three datasets and directly follow the evaluation protocol of RT-Gene.
More concretely, after data pre-processing, MPIIFaceGaze contains 45K images of 15 subjects. We perform leave-one-person-out evaluation on it.
EyeDiap contains 16K images of 14 subjects. 
We perform four-folder cross validation on it.
Gaze360 contains 84K images of 54 subjects for training and 16K images of 15 subjects for test.
RT-Gene contains 92K images of 13 subjects. We follow the evaluation protocol and split the dataset into three subsets. A three-folder cross validation is performed in RT-Gene.

\textbf{Training.}
\label{sec:traninig} 
We train GazeTR-Pure and GazeTR-Hybrid in ETH-XGaze with 512 batch sizes and 50 epochs.
The learning rate is set as 0.0005 and 0.5 weight decay.
The decay steps is set as 20 epochs for GazeTR-Pure and 10 epochs for GazeTR-Hybrid.
Adam optimizer~\cite{kingma2014adam} is used to train the two models with $\beta_1=0.9$ and $\beta_2=0.999$. 
We use a linear learning rate warmup, which is set as 5 epochs in GazeTR-Pure and 3 epochs in GazeTR-Hybrid.

For the four evaluation datasets, most of learning parameters are inherited from the above setting. 
In addition, we set batch sizes as 512, epochs as 80, decay steps as 60 and epochs of warmup as 5 in MPIIFaceGaze for GazeTR-Hybrid.
The four parameters are set as (16, 80, 60, 5) in EyeDiap,  (256, 80, 60, 5) in Gaze360 and (512, 80, 60, 5) in RT-Gene.
GazeTR-Pure keeps the same setting as GazeTR-Hybrid with changing the decay steps to 40.
The two models both are implemented using PyTorch and trained on NVIDIA Tesla V100 GPUs.

\textbf{Evaluation metrics.}
We use angular error as the evaluation metric as most of gaze direction estimation methods. A smaller error represents a better model. 

\begin{table}[t]
	\setlength\tabcolsep{8pt}
	\renewcommand\arraystretch{1.3}
	\centering
	\normalsize
	\caption{Comparison with State-of-the-art methods. GazeTR-Pure cannot achieve competitive results while GazeTR-Hybrid shows state-of-the-art results in all datasets.}
	\begin{threeparttable}

		\begin{tabular}{ccccc}
			\toprule
			\textbf{Methods}&MPIIFaceGaze~\cite{Zhang_2017_CVPRW}&EyeDiap~\cite{Mora_2014_ETRA}&Gaze360~\cite{Kellnhofer_2019_ICCV}&RT-Gene~\cite{Fischer_2018_ECCV}\\
	
			\midrule
			
			FullFace~\cite{Zhang_2017_CVPRW}&$4.93^\circ$&$6.53^\circ$&$14.99^\circ$&$10.00^\circ$\\
			RT-Gene~\cite{Fischer_2018_ECCV}&$4.66^\circ$&$6.02^\circ$&$12.26^\circ$&$8.60~^\circ$\\
			Dilated-Net~\cite{Chen_2019_ACCV}&$4.42^\circ$&$6.19^\circ$&$13.73^\circ$&$8.38~^\circ$\\
			Gaze360~\cite{Kellnhofer_2019_ICCV}&$4.06^\circ$&$5.36^\circ$&$11.04^\circ$&$7.06^\circ$\\
			CA-Net~\cite{Cheng_2020_AAAI}&$4.27^\circ$&$5.27^\circ$&$11.20^\circ$&$8.27^\circ$\\
			\midrule
			Mnist~\cite{Zhang_2015_CVPR}&$6.39^\circ$&$7.37^\circ$&\textit{N/A}&$14.9^\circ$\\
			GazeNet~\cite{Zhang_2017_tpami}&$5.76^\circ$&$6.79^\circ$&\textit{N/A}& \textit{N/A}\\
			
			\midrule
			GazeTR-Pure & $4.74^\circ$&$5.72^\circ$&$13.58^\circ$&$8.06^\circ$\\
			GazeTR-Hybrid &$\bm{4.00^\circ}$  & $\bm{5.17^\circ}$  & $\bm{10.62^\circ}$ &$\bm{6.55^\circ}$ \\
			\bottomrule
		\end{tabular}
		
\end{threeparttable}

\label{tab:3dgaze}
\end{table}

\begin{table}[t]
	\setlength\tabcolsep{8pt}
	\renewcommand\arraystretch{1.3}
	\centering
	\normalsize
	\caption{Comparison with the State-of-the-art POG estimation method. We transfer the GazeTR-Hybrid to perform POG estimation. The accuracy of gaze direction is converted from POG estimation result with post-processing. The result shows GazeTR-Hybrid perform better than AFF-Net.}
	\begin{threeparttable}
		
		\begin{tabular}{ccccc}
			\toprule
			\multirow{2}{*}{\textbf{Methods}}&\multicolumn{2}{c}{MPIIFaceGaze~\cite{Zhang_2017_CVPRW}}&\multicolumn{2}{c}{EyeDiap~\cite{Mora_2014_ETRA}}\\
			&POG&Gaze Direction&POG&Gaze Direction.\\
			\midrule
			AFF-Net~\cite{Bao_2020_ICPR}&$4.21$ cm & $3.73^\circ$& $9.25$ cm    &$6.41^\circ$\\	
			GazeTR-Hybrid (POG) &$\bm{3.6}$ cm & $\bm{3.08^\circ}$  &$\bm{7.81}$ cm & $\bm{5.52^\circ}$  \\
			\bottomrule
		\end{tabular}
		
	\end{threeparttable}
	
	\label{tab:2dgaze}
\end{table}

\subsection{Comparison with State of The Art }
We first compare GazeTR-Pure and GazeTR-Hybrid with state-of-the-art methods.
The result is shown in~\Tref{tab:3dgaze}.
Note that, Mnist~\cite{Zhang_2015_CVPR} and GazeNet~\cite{Zhang_2017_tpami} estimate the gaze direction originated from eye centers, which is different from ours.
We use post-processing method~\cite{cheng2021appearance} to convert their results for fair comparison.
\Tref{tab:3dgaze} shows  GazeTR-Pure cannot achieve competitive results while GazeTR-Hybrid shows state-of-the-art results in all benchmarks.

On the other hand, we notice~\cite{cheng2021appearance} reports AFF-Net can achieve better result in MPIIFaceGaze than GazeTR-Hybrid via point-of-gaze (POG) estimation.
We also transfer the GazeTR-Hybrid to  point-of-gaze (POG) estimation.
More concretely, we change the output of GazeTR-Hybrid to perform POG estimation and then convert results into gaze directions with post-processing, which is the same as AFF-Net.
Note that, we slightly modify the GazeTR-Hybrid as GazeTR-Hybrid (POG) with adding the information of eye and face corner positions into the last MLP as AFF-Net.
The GazeTR-Hybrid (POG) directly inherits the pre-trained parameters of GazeTR-Hybrid. It is not pre-trained in ETH-XGaze again.
The results is shown in ~\Tref{tab:2dgaze}.
Not only POG accuracy but also converted gaze direction accuracy are shown in the table.
The evaluation metric of POG estimation is euclidean distance.
Our method significantly outperforms AFF-Net.

\begin{table}[t]
	\centering
	\renewcommand\arraystretch{1.2}
	\setlength\tabcolsep{3pt}
	\normalsize
	\caption{Evaluation of hyper-parameters in the hybrid transformer.}

	\begin{tabular}{llcccc}
		\toprule
		\multicolumn{2}{c}{Hyper-Parameters}&MPIIFaceGaze~\cite{Zhang_2017_CVPRW}&EyeDiap~\cite{Mora_2014_ETRA}&Gaze360~\cite{Kellnhofer_2019_ICCV}&RT-Gene~\cite{Fischer_2018_ECCV} \\
		\midrule

		\multirow{4}{*}{Layers $L$ } & $L=6$  &$4.00^\circ$ & $5.17^\circ$  &$10.62^\circ$&$6.55^\circ$\\
		&$L=12$ &$3.99^\circ$&$5.29^\circ$&$10.82^\circ$&$6.84^\circ$\\
		&$L=24$ &$3.96^\circ$& $5.31^\circ$&$10.73^\circ$ &$6.64^\circ$\\
		&$L=36$ &$3.97^\circ$&$5.28^\circ$&$10.64^\circ$&$6.53^\circ$\\
		\midrule
		\multirow{4}{*}{Heads $N$ } & $N=1$ &$4.04^\circ$ & $5.09^\circ$  & $10.64^\circ$& $6.86^\circ$\\
		&$N=4$ & $3.88^\circ$ & $5.28^\circ$ & $10.56^\circ$ & $6.57^\circ$\\
		&$N=8$ & $4.00^\circ$ & $5.17^\circ$ & $10.62^\circ$ & $6.55^\circ$\\
		&$N=16$ & $3.90^\circ$ & $5.30^\circ$ & $10.69^\circ$ & $6.74^\circ$\\
		\midrule
		\multirow{3}{*}{Input dimensions $D$} & $D=32$ &$4.00^\circ$ & $5.17^\circ$  &$10.62^\circ$& $6.55^\circ$\\
		&$D=256$ &$3.93^\circ$&$5.25^\circ$&$10.59^\circ$&$6.57^\circ$\\
		&$D=512$ &$4.07^\circ$& $5.42^\circ$&$11.20^\circ$ &$6.73^\circ$\\
		\bottomrule
	\end{tabular}
	\label{table:hyper}

\end{table}

\subsection{Hyper-parameters in Hybrid Transformer}
Transformers have some adjustable hyper-parameters.
In this section, we conduct experiment to evaluate these hyper-parameters.
Overall, we evaluate three hyper-parameters in GazeTR-Hybrid, which are the number of layers in transformer, the number of heads in multi-head self-attention and the input dimension of transformer.

\textbf{Layers.} The transformer can be easily stacked as a multi-layer transformer which is similar to CNNs.
We first evaluate the effect of depth in multi-layers transformer.
As shown in the second row of~\Tref{table:hyper}, we evaluate the performance when the number $L$ of layers is $6$, $12$,  $24$ and $36$.  
It is interesting that the performance is decreased when $L$ is increased to 12.
Then, the performance is slightly improved with further increasing $L$. 
The 36-layer model outperforms 6-layer model in MPIIFaceGaze and RT-Gene datasets while also has worse performance than 6-layer model in EyeDiap and Gaze360 datasets. Note that, the $L$ only represents the number of layers in transformers. They all contain an additional convolutional layers of ResNet-18 for feature extraction. 

\textbf{Heads.} The multi-head self-attention layer is important in transformers.
The layer uses multiple heads to project the feature into several different subspaces.
We further evaluate the impact of the number $N$ of heads.
The result is shown in the third row of \Tref{table:hyper}.
Experiment shows more heads cannot bring performance improvement.
$N=1$ has best result in EyeDiap while cannot perform well in other three datasets.
$N=4$ has best results in MPIIFaceGaze and Gaze360. It also has acceptable results in EyeDiap and RT-Gene.
Although $N=16$ has good performance in MPIIFaceGaze, they have the worst results in EyeDiap and Gaze360.

\textbf{Input dimensions.}
GazeTR-Hybrid uses an $1\times 1$ convolutional layer to control the input dimension of transformers,~\ie, the channels of feature maps.
We evaluate the impact of the size $D$ of the channels.
As shown in the fourth row of ~\Tref{table:hyper}, slightly increasing $D$ into 256 brings improvement in MPIIFaceGaze and Gaze360.
However, further increasing $D$ into 512 degrade performance.

\vspace{-1.5mm}
\subsection{Ablation Study}
\vspace{-1.5mm}
We conduct ablation study in this section. 
We respectively conduct two ablation studies for the transformer and the convolutional layers.
We ablate the self-attention mechanism in the transformer and decrease some convolutional layers.

\begin{table}[t]
	\centering
	\renewcommand\arraystretch{1.2}
	\setlength\tabcolsep{2pt}
	\normalsize
	\caption{Ablation study. Self-attention brings significant performance improvement. Deep convolution is also necessary for GazeTR-Hybrid. }
	\begin{tabular}{lcccc}
		\toprule
		Methods &MPIIFaceGaze~\cite{Zhang_2017_CVPRW}&EyeDiap~\cite{Mora_2014_ETRA}&Gaze360~\cite{Kellnhofer_2019_ICCV}&RT-Gene~\cite{Fischer_2018_ECCV} \\
		
		\midrule
		GazeTR-Hybrid ($L=6$) & $4.00^\circ$ & $5.17^\circ$  & $10.62^\circ$ & $6.55^\circ$\\
		\textit{w/o} self-attention & $4.19^\circ$ & $5.43^\circ$ & $10.96^\circ$ & $6.56^\circ$\\
		\textit{w/o} deep convolution (6-layer conv.)&$4.45^\circ$&$5.50^\circ$&$12.38^\circ$ & $7.51^\circ$\\
		\midrule
		GazeTR-Hybrid ($L=12$) & $3.99^\circ$ & $5.29^\circ$ & $10.82^\circ$ & $6.84^\circ$\\
		\textit{w/o} self-attention &$4.02^\circ$&$5.50^\circ$ &$11.12^\circ$&$6.68^\circ$\\
		\textit{w/o} deep convolution (6-layer conv.)&$4.59^\circ$&$5.30^\circ$&$12.53^\circ$ & $7.52^\circ$\\
		\midrule
		GazeTR-Hybrid ($L=24$) &$3.96^\circ$& $5.31^\circ$&$10.73^\circ$ &$6.64^\circ$\\
		\textit{w/o} self-attention&$3.98^\circ$&$5.45^\circ$ &$11.07^\circ$&$6.64^\circ$\\
		\textit{w/o} deep convolution (6-layer conv.) &$4.62^\circ$&$5.34^\circ$&$12.55^\circ$& $7.68^\circ$\\
		\midrule
		GazeTR-Hybrid ($L=36$) &$3.97^\circ$&$5.28^\circ$&$10.64^\circ$&$6.53^\circ$\\
		\textit{w/o} self-attention &$4.01^\circ$&$5.67^\circ$&$11.53^\circ$&$7.32^\circ$\\
		\textit{w/o} deep convolution (6-layer conv.) &$4.57^\circ$&$5.38^\circ$&$12.51^\circ$&$7.95^\circ$\\
		\bottomrule
	\end{tabular}
	\label{table:ablation}
\end{table}

\textbf{\textit{w/o} self-attention.} Self-attention is the core of transformers.
It adaptively learns attention weights for each feature. 
To evaluate the impact of self-attention, we fix the learned attention weights as average weights and do not change the architecture of GazeTR-Hybrid. The result is shown in~\Tref{table:ablation}. We ablate the self-attentionin four hybrid transformers for comprehensive comparison, where the $L$ is the number of layers in transformers.
\Tref{table:ablation} proves self-attention is effective in gaze estimation tasks.
The largest improvement is $0.99^\circ$ in the experiment of GazeTR-Hybrid ($L=36$) in Gaze360.

\textbf{\textit{w/o} deep convolution.} On the other hand, we explore the impact of convolutional layers in GazeTR-Hybrid.
The GazeTR-Hybrid use the convolutional layers of ResNet-18.
It contains a total of $17$ convolutional layers. We remove most of convolutional layers and only preserve the first $5$ convolutional layers. An additional $1\times 1$ convolutional layers (strides$=2$) is added after the third convolutional layer for downsample.
We finally acquire feature maps $\in\mathbb{R}^{28\times 28 \times 64}$.
We also split the feature maps into $7 \times 7$ patches where each patch $\in\mathbb{R}^{4*4*64}$. To ensure the same input dimensions of transformers with GazeTR-Hybrid, we linearly project each patch into a $32$ feature vector. 
The result is shown in~\Tref{table:ablation}.
\textit{w/o} deep convolution brings obvious performance degradation.
This indicates the necessary of convolutional layers.
In practice, this experiment also can be regarded as pure transformers plus light weight CNNs.  

\begin{table}[t]
	\centering
	\renewcommand\arraystretch{1.2}
	\setlength\tabcolsep{5pt}
	\normalsize
	\caption{Comparison with Pure CNNs. GazeTR-Hybrid significantly outperforms GazeTR-Conv with an additional transformer. It also performs better than ResNet-50 with less parameters.}
	\begin{tabular}{l|c|cccc}
		\toprule
		Methods & Params &MPIIFaceGaze~\cite{Zhang_2017_CVPRW}&EyeDiap~\cite{Mora_2014_ETRA}&Gaze360~\cite{Kellnhofer_2019_ICCV}&RT-Gene~\cite{Fischer_2018_ECCV} \\
		\midrule
		
		GazeTR-Pure& $104$ M & $4.74^\circ$&$5.72^\circ$&$13.58^\circ$&$8.06^\circ$\\
		GazeTR-Hybrid& $11.4$ M  &$4.00^\circ$ &$5.17^\circ$  &$10.62^\circ$&$6.55^\circ$\\
		\midrule
		GazeTR-Conv& $11.2$ M &$4.21^\circ$&$5.35^\circ$&$11.09^\circ$&$6.95^\circ$\\
		ResNet-50& $25.5$ M &$4.11^\circ$&$5.27^\circ$&$10.73^\circ$&$6.54^\circ$\\
		\bottomrule
	\end{tabular}
	\label{table:purecnn}
\end{table}

\subsection{Comparison with Pure CNNs.}
The GazeTR-Hybrid integrates CNNs and transformers to achieve high performance.
In this section, we conduct the experiment to prove the advantage of GazeTR-Hybrid compared with pure CNNs.

We take the convolutional layers of GazeTR-Hybrid out and add an additional two-layer MLP after  the convolutional layers for gaze estimation.  
The hidden layer of the MLP is set as $256$.
We refer the new network as GazeTR-Conv.
In practice, GazeTR-Conv has a similar architecture as ResNet-18.
On the other hand, we select the popular ResNet-50 as another compared method.
The two methods are both first pre-trained in ETH-XGaze and then trained in other datasets for evaluation. 
We summarize the result in~\Tref{table:purecnn}.
We also count the parameters of these models and show it in the second column. 
The performance of GazeTR-Pure is also added into~\Tref{table:purecnn} for reference. 

As shown in~\Tref{table:purecnn}, GazeTR-Hybrid brings significant improvement compared with GazeTR-Conv.
This proves the effectiveness of integrating CNNs and transformers.
More important, the transformer only brings extra $0.2$M parameter costs.
On the other hand, ResNet-50 doubles the parameters of GazeTR-Conv.
It does bring performance improvement compared with GazeTR-Conv.
However, GazeTR-Hybrid also achieves higher performance than ResNet-50 with less parameters.
\begin{figure}[t]
	\begin{center}
		\includegraphics[width=\columnwidth]{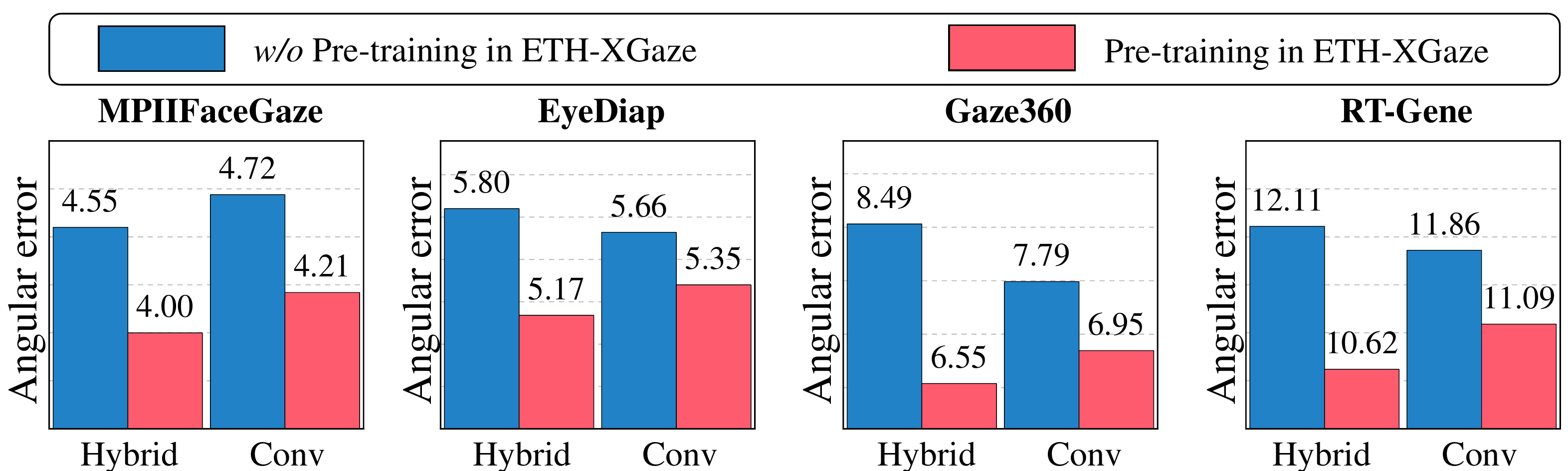}	
	\end{center}
	\caption{The impact of pre-training for transformers. We show the performance of Hybrid (GazeTR-Hybrid) and Conv (GazeTR-Conv) without pre-training. It is obvious that pre-training brings significant improvement. Meanwhile, the transformer can bring larger improvement with pre-training. }
	\label{fig:pretrain}
\end{figure}

\subsection{Impact of Pre-training} 

The researches on other tasks indicate transformers require a large-scale dataset for pre-training~\cite{dosovitskiy2020image}.
We also provide the result of GazeTR-Hybrid without pre-training for deeper understanding.
We directly evaluate the GazeTR-Hybrid in four evaluation datasets without pre-training.
Meanwhile, to show the impact of transformers, we also evaluate the GazeTR-Conv for comparison. We show the result in~\Fref{fig:pretrain}.
According the result, it is easy to have the following conclusion:
\vspace{-2mm}
\begin{itemize}
	\setlength{\itemsep}{3pt}
	\setlength{\parsep}{0pt}
	\setlength{\parskip}{0pt}
	\item[-]  Pre-training is useful for both GazeTR-Hybrid and GazeTR-Conv. It brings significant improvement in all evaluation datasets.
	
	\item[-] Pre-training is more important for transformers. Without pre-training, GazeTR-Hybrid performs worse than GazeTR-Conv in three datasets. However, GazeTR-Hybrid shows better performance than GazeTR-Conv in all datasets with pre-training.  
\end{itemize}

\subsection{Performance Improvement of Self-attention.}

\begin{figure}[t]
	\begin{center}
		\includegraphics[width=0.95\columnwidth]{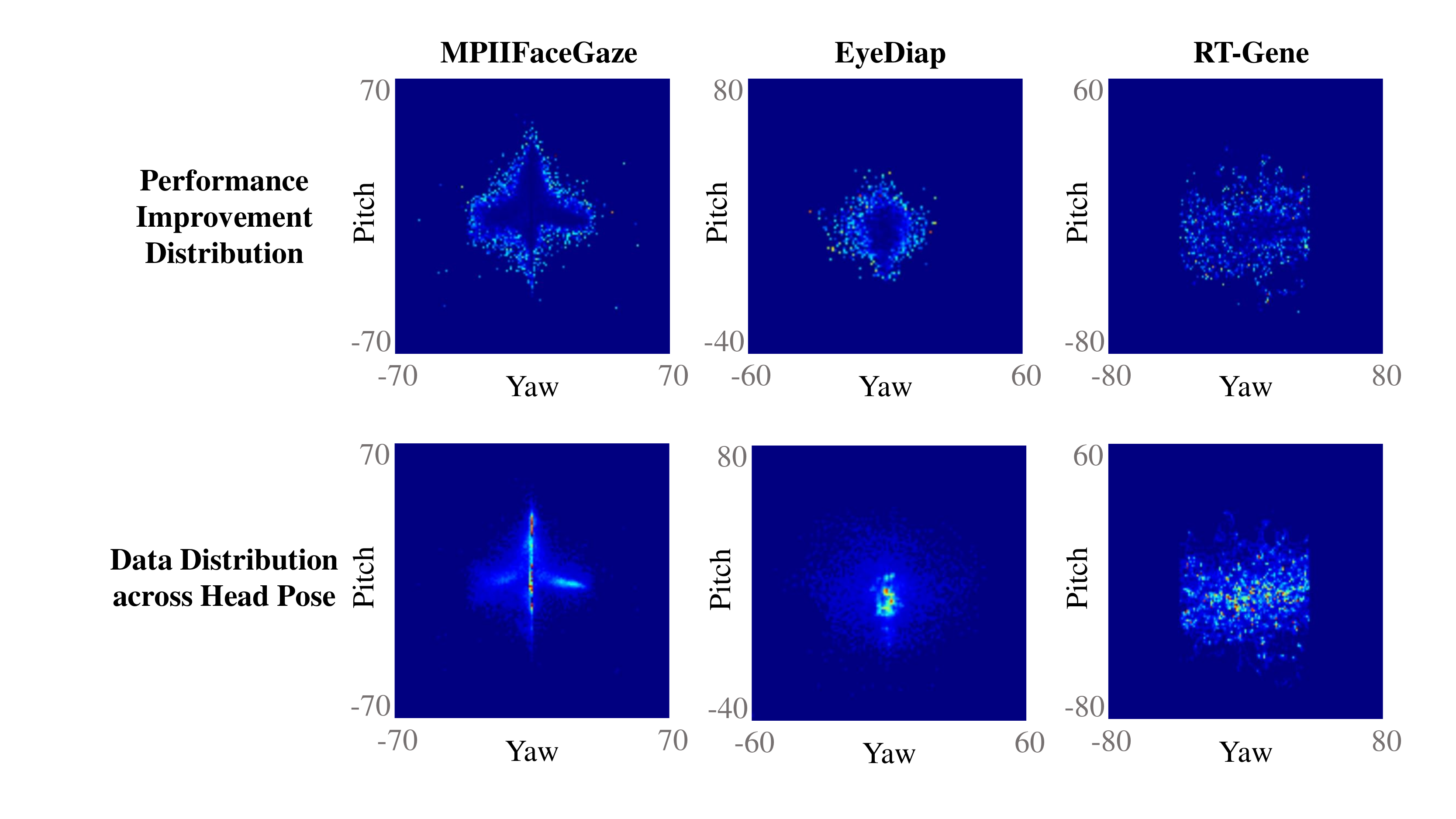}	
	\end{center}
	\caption{The performance improvement of self-attention. The first row shows the performance improvement distribution of GazeTR-Hybrid compared with GazeTR-Hybrid (\textit{w/o} self attention). The second row is the head pose distribution of whole datasets. The x-axis and y-axis in each figure is the yaw and pitch of head pose. A brighter point represents larger improvement or more samples. It is interesting that outer regions have larger improvement than central regions.}
	\label{fig:distribution}
\end{figure}

We further analyze the outputs of GazeTR-Hybrid and GazeTR-Hybrid (\textit{w/o} self attention) to understand the advantage of self-attention .
We visualize the performance improvement of self-attention to GazeTR-Hybrid.
We cluster samples according to their head pose where each cluster is a $1^\circ\times 1^\circ$ local  region.
We count the average performance changes in each cluster, and remove the region which has performance degradation.
The result is shown in~\Fref{fig:distribution}.
The brighter the point, the larger the improvement. 
We also show the data distribution of whole datasets for reference.
The brightness of points is decided by the number of samples in clusters.
It is interesting that the outer regions have large improvement and the central regions have relative small improvement.

\subsection{Discussion}
\textbf{\textit{1). Transformer in gaze estimation.}} We follow popular ViT-Base to design the GazeTR-Pure in this paper. The GazeTR-Pure has $104$M parameters and cannot achieve competitive results compared with GazeTR-Hybrid which has $11.4$M parameters. However, we do not believe the result indicates pure transformers is unsuited to gaze estimation. On the other hand, we conduct experiment to demonstrate the effectiveness of self-attention.
This indicates the potential of transformer in gaze estimation.

\textbf{\textit{2) Pre-training in gaze estimation.}}
Previous gaze estimation methods usually use off-the-shelf networks pre-trained in ImageNet~\cite{deng2009imagenet} as backbone~\cite{Zhang_2017_CVPRW,Kellnhofer_2019_ICCV}.  
In this paper, we propose to pre-train models in large-scale gaze datasets.
We believe: 1) Technology development makes gaze estimation models become more complex.
Pre-training rather than training from scratch is necessary for future methods. 
2) Recently, many large-scale gaze datasets have been proposed. It is possible that future methods rely on large-scale gaze datasets rather than the ImageNet for pre-training. In \cite{ranjan2018light}, the author also proves different pre-training tasks brings large accuracy difference in gaze estimation.

\section{Conclusion}
In this paper, we explore the performance of transformers in gaze estimation.
We consider two forms of transformers, which are pure transformers and hybrid transformers.
Experiments show hybrid transformers perform better than pure transforms, and can achieve state-of-the-art results in four popular datasets with pre-training.
We also further conduct experiments to assess the hybrid transformers.
This paper provides a new direction of future gaze research.

\bibliography{gaze}
\end{document}